\documentclass[runningheads]{llncs}
\usepackage[T1]{fontenc}
\usepackage{graphicx}
\usepackage{hyperref}

\usepackage{color}

\begin{document}
\title{ChartFormer: A Large Vision Language Model for Converting Chart Images into Tactile Accessible SVGs}
\titlerunning{ChartFormer}



%
\author{Omar Moured\inst{1,2}\orcidID{0000-0003-4227-8417}
\and
Sara Alzalabny\inst{3}\orcidID{0000-0002-3533-2473}
\and
Anas Osman\inst{4}\orcidID{0009-0000-9938-1226}
\and
Thorsten Schwarz\inst{2}\orcidID{0000-0002-7346-5744}
\and
Karin Müller\inst{2}\orcidID{0000-0003-4309-1822}
\and
Rainer Stiefelhagen\inst{1,2}\orcidID{0000-0001-8046-4945}
}

\authorrunning{O. Moured et al.}
%
\institute{
CV:HCI@KIT, Karlsruhe Institute of Technology, Germany\\
\and
ACCESS@KIT, Karlsruhe Institute of Technology, Germany\\
\and
NeptunLab, University of Freiburg, Germany\\
\and
Middle East Technical University, Turkey
}

\maketitle
\begin{abstract}
Visualizations, such as charts, are crucial for interpreting complex data. However, they are often provided as raster images, which are not compatible with assistive technologies for people with blindness and visual impairments, such as embossed papers or tactile displays. At the same time, creating accessible vector graphics requires a skilled sighted person and is time-intensive. In this work, we leverage advancements in the field of chart analysis to generate tactile charts in an end-to-end manner. Our three key contributions are as follows: (1) introducing the \textit{ChartFormer} model trained to convert raster chart images into tactile-accessible SVGs, (2) training this model on the \textit{Chart2Tactile} dataset, a synthetic chart dataset we created following accessibility standards, and (3) evaluating the effectiveness of our SVGs through a pilot user study with an refreshable two-dimensional tactile display. Our work is publicly available at \url{https://github.com/nsothman/ChartFormer}.

\end{abstract}
\section{Introduction}
Charts are essential for communicating complex information to a broad and diverse audience. People with blindness and visual impairments access these charts primarily through text-based descriptions (e.g., screen readers), and tactile representations (e.g., embossed paper, tactile displays). The use of tactile charts in educational setting has been shown to enhance the development of people with blindness and visual impairments' analytical skills \cite{paters}. These tactile charts are often created using drawing software for vector graphics (e.g., Inkscape, LibreOffice Draw, etc.) and saved in the Scalable Vector Graphics (SVG) format.
SVGs are XML-based files that store geometrical shapes using mathematical formulas in a hierarchical structure. This format presents several advantages for creating accessible graphics \cite{Bernhard}, including (1) each element in an SVG can be assigned different styles, translating into distinctive textures in the tactile version. (2) SVG files can hold supplementary textual descriptions which enhances interactivity when used with screen readers or tactile displays. (3) ability to be resized without causing blurring or distortion, making them ideal for varying paper sizes or zooming on tactile displays \cite{moured2023accessible}.
Creating an SVG chart from a raster image requires \textit{careful simplification of both textual and visual content to support tactile formats while also preserving the integrity of the chart information}. This includes reducing textual content, such as limiting the number of axis labels and focusing on only crucial visual elements, like emphasizing significant scatter points in a scatter plot. Due to these complexities, crafting vector graphics is not a trivial task. Nevertheless, AI-based chart analysis models have demonstrated emerging capabilities in image analysis. In light of these developments, we outline three main contributions in our work: (1) We introduce a transformer-based model that extracts key information and assigns styles for the SVG file. (2) We present the Chart2Tactile dataset, created in adherence to accessibility guidelines. (3) We share insights from a pilot user study involving four blind participants who evaluated SVG graphics generated by our model on a 2D tactile display.
\section{Related Work}
Very few studies have explored automated heuristics for making raster charts tactically accessible. Most prior research follows a two-step methodology to convert charts, beginning with metadata extraction and followed by conversion into tactile format. In this section, we discuss these methods and also the available benchmarks. 

\subsection{Metadata Extraction}
A recent work, ChartLLama \cite{han2023chartllama}, trained a large Vision-Language Model (VLM) on synthetically generated images across 10 chart types. One of its tasks is converting charts into Matplotlib Python codes. While this model demonstrates the potential of VLMs in chart comprehension, it overlooks accessibility concerns. ChartDetective \cite{chartdetective} provides a UI for processing SVG chart documents, allowing data extraction through simple drag-and-drop interactions. This tool is useful when SVGs are available, but additional steps are still required for redrawing the data into accessible tactile materials.

\subsection{Tactile Charts}
A limited number of studies have proposed systems capable of both extracting data and generating tactile materials. Engel et al. presented the \textit{SVGPlott} system \cite{SVGPlott}, a GUI that comprises six steps to convert bitmap images into printable SVG format. This system streamlines the process through step-by-step guidance; however, it still requires manual extraction of metadata, which may not be feasible for dense charts. A more recent development, Chart4Blind \cite{moured2024chart4blind}, introduced an end-to-end AI-assisted user interface designed to convert charts into tactile formats accessible for the visually impaired. Nevertheless, it currently only supports line charts, highlighting the need for broader support of other types of visualizations.

\subsection{Available Datasets}
ChartAssistants \cite{meng2024chartassisstant}, a recent work, stands out for its comprehensive collection of chart images, each paired with detailed metadata. Although it lays a solid groundwork for converting charts to code, it cannot generate accessible visualizations. In contrast, datasets oriented towards accessibility, like VisText \cite{tang2023vistext}, have focused on making visualizations accessible through chart summarization tasks, but none have considered the tactile modality. \textbf{To our knowledge, we propose the first dataset for the task of \textit{Chart2Tactile} conversion.}
\begin{figure}
    \centering
    \includegraphics[width=\textwidth]{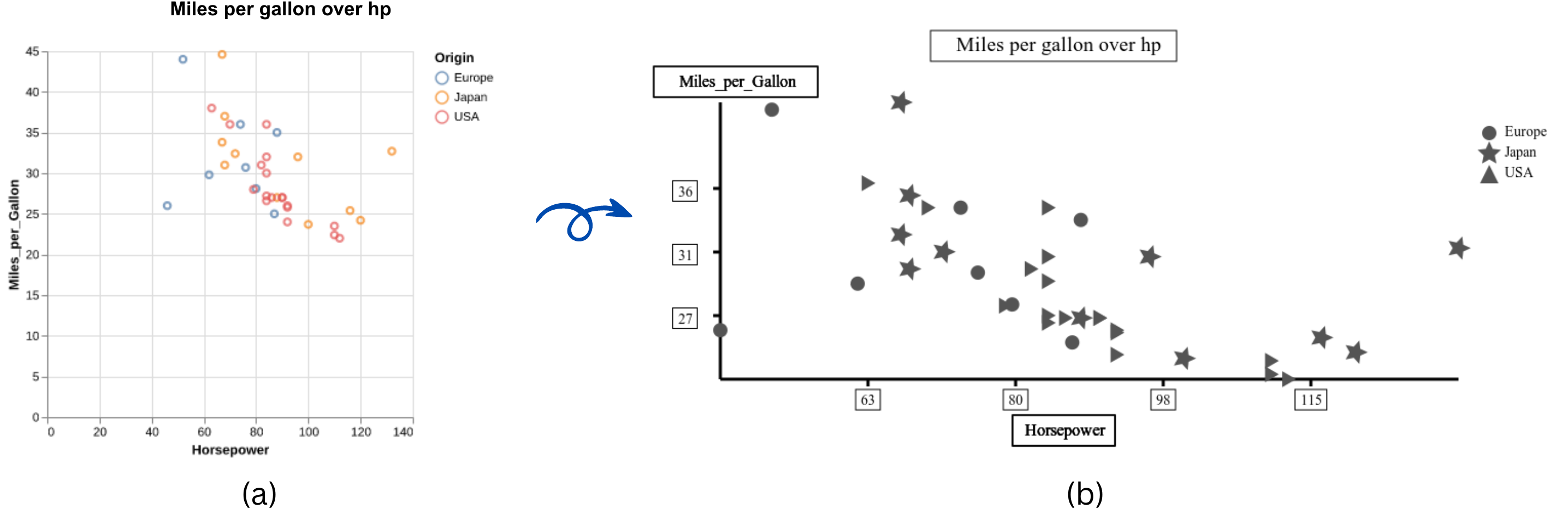}
    \caption{A scatter plot sample: (a) the original synthesized raster image; (b) the tactile version following accessibility guidelines.}
    \label{fig:sampledata}
\end{figure}

\section{Chart2Tactile Dataset} 
Our dataset comprises 10,000 tactile chart images, spanning 4 categories (line, bar, scatter and error-bar charts) each accompanied by time series data and a raster version. Below, we describe the formation of the \textit{Chart2Tactile} dataset.

\subsection{Metadata Collection} 
To create a visualization, time-series data is essential. We aimed to select a source that would be augmented by our contributions. We identified the VisText \cite{tang2023vistext} and ChartX \cite{xia2024chartx} datasets as the most suitable choices. VisText offers 8,822 images, complete with their data tables and accessible summarizations, featuring univariate time series. ChartX contains 48K chart data covering 22 topics, 18 chart types, with each chart including four modalities: image, CSV, Python code, and text description. A sample raster image is presented in Figure \ref{fig:sampledata}-(a).

\subsection{SVG Guidelines}
\label{svgguidlines}
Rendering the metadata as tactile charts necessitates adherence to established guidelines to ensure that the charts are accessible by individuals with visual impairments. This process involves not only the translation of visual information into a tactile format but also the thoughtful consideration of how various elements can be differentiated by touch. We followed various tactile printing guidelines \cite{BANA2010,SVGPlott} to create accessible SVGs. The key requirements we adhered to are summarized as follows:

\begin{enumerate}
    \item Elements should be distinguishable by touch, using varying thicknesses or symbol types such as dotted or dashed patterns.
    \item Thin elements should be avoided.
    \item Text in tactile illustrations should be in Braille, oriented horizontally.
\end{enumerate}

Additionally, we collaborated with an expert from the Center for Digital Accessibility and Assistive Technology\footnote{\url{https://www.access.kit.edu/english/index.php}} at Karlsruhe Institue of Technology, specializing in converting educational materials for people with blindness and visual impairments. Their feedback included the following recommendations:

\begin{enumerate}
    \item Enclose text content with a bounding box for better exploration and distinguishing separate texts more effectively.
    \item For dense charts such as scatter plots, only significant, non-overlapping points should be drawn to avoid clutter.
    \item Embed description tags for both text and visual elements to enable accessibility via screen readers.
\end{enumerate}

\subsection{SVG Creation}
For transforming metadata into SVGs, we used the \textit{svgwrite} Python package\footnote{\url{https://svgwrite.readthedocs.io/en/latest/}}. For each time series, we synthesized an SVG template and rendered a raster image using \textit{Vega-Lite}\footnote{\url{https://vega.github.io/vega-lite/}}. To ensure accuracy, we manually selected samples from each category of the data and conducted a thorough verification process. A tactile sample is illustrated in Figure \ref{fig:sampledata}-(b).
\begin{figure}
    \centering
    \includegraphics[width=0.97\textwidth]{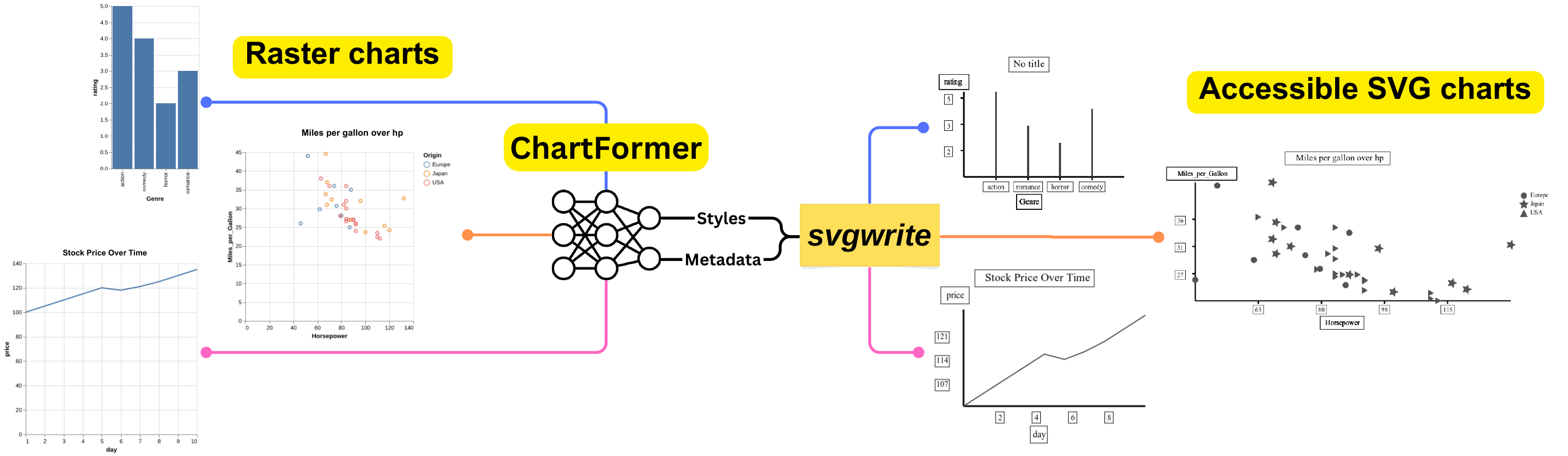}
    \caption{The ChartFormer takes a raster x-y plot as an input. The essential metadata and styles are extracted, which are then used to populate the svgwrite templates. For better viewing resolution, please visit our project page.}
    \label{fig:system}
\end{figure}

\section{ChartFormer Model}
We selected LLaVA-1.5 \cite{liu2023improved} as our baseline model, comprising a vision encoder for image input and a language model for text output decoding. We used the baseline weights from ChartLLama \cite{han2023chartllama}, adopted the same hyperparameters for training, and then fine-tuned the model for 10 epochs using our dataset. The model is trained to analyze x-y raster chart images as input and extract the simplified metadata with the styling for the \textit{svgwrite} code (see Figure \ref{fig:system}). More specifically, the following information is extracted:
\begin{enumerate}
    \item The x-y chart type, and titles, including plot, axes and legend titles.
    \item Axes range as 3 or 4 labels covering the whole period. The labels should adhere to the encodings (e.g., int, float, fraction, date/time and text). 
    \item Extract the time-series data for drawing.
\end{enumerate}

The extracted data are then rendered using predefined svgwrite code templates for each chart category. It is important to note that for the scatter plot, we draw 10 points per label unit while separating overlapping points, in adherence to the SVG Guidelines \ref{svgguidlines}.

\section{Pilot User Study}
We conducted a pilot user study with four people with blindness and visual impairments (three males and one female) to evaluate the effectiveness of the generated SVGs on a HyperBraille 2D tactile display \footnote{https://metec-ag.de/en/produkte-graphik-display.php}.

\subsection{Procedure}
\begin{figure}[h]
    \centering
    \includegraphics[width=\textwidth]{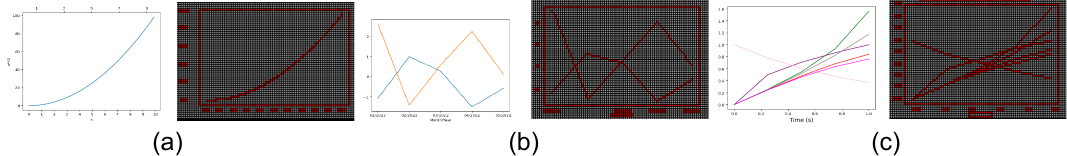}
    \caption{SVG-formatted line charts used in the user study, showcasing varying complexities: (A) a single line; (B) two lines; (C) six lines. For better viewing resolution, please visit our project page.}
    \label{fig:charts}
\end{figure}

The user study images were randomly sampled from the LG dataset \cite{moured2023line}, which includes real charts. At the beginning of the session, a test graph was provided to introduce the participants to the available interactions. Afterwards, the $3$ line charts shown in Figure \ref{fig:charts} were displayed and the participants were asked to explain and identify the key elements, titles, labels and legends and count lines, in each graph, as well as name few points intersection and line trend.

\subsection{Results \& Discussion}
All participants successfully completed tasks related to charts (A) and (B), which involved identifying intersections and counting lines. They could also accurately describe the line trend as increasing, decreasing, or constant. However, in chart (C), participants encountered difficulties in counting all intersections, likely due to the chart's high density. Two participants used zoom features on the tactile display to discern closely positioned elements and intersections more clearly. They also appreciated the audio descriptions, which facilitated access to the chart's textual elements. A common suggestion from all participants was regarding SVG rendering, specifically to address the staircasing effect in the tactile output. The need for smoother line rendering to avoid jagged or stair-like appearances was emphasized. 

\subsection{Limitations}
Although our system has been positively received by the participating people with blindness and visual impairments and collaborators, we believe there is still significant room for improvement: (1) Our system mainly targets x-y plots with two axes and charts of a single type. Future implementations could encompass other chart types. (2) Adding an interface to our system could allow sighted individuals to modify the chart before exporting it, ensuring that textual and visual details are accurately represented. (3) Conducting a larger, formal user study is necessary to assess the performance and furthermore, to experiment with different types of charts beyond just line charts.
\section{Conclusion}
In this work, we have showcased the potential of vision-language models for enhancing accessibility through the creation of tactile graphics and descriptive content. Current VL models fail to meet the special requirements that different people may need, such as those with visual impairments. It is important for future research to address it by refining models to better serve these diverse groups. We believe that available models such as ChatGPT or Gemini can be utilized to address these tasks with few- or zero-shot tuning, hence minimal computational demands, offering an easier available solution for schools and higher education institutes. We have also introduced the first dataset for tactile visualizations that complies with accessibility guidelines. Despite some limitations, our contributions aim to encourage further research into supporting accessible graphics production.
\newpage
\begin{credits}
\subsubsection{\ackname}
The authors would like to thank the HoreKa computing cluster at KIT for the computing resources used to conduct this research.

\subsubsection{\discintname}
This research was funded by the European Research Council (ERC) grant and the European Union’s Horizon $2020$ research and innovation program under the Marie Sklodowska-Curie grant agreements no. $816006$ and $861166$ respectively. 

\end{credits}

%
%
%
\bibliographystyle{splncs04}
\bibliography{mybib}
\end{document}